\documentclass[10pt]{IEEEtran}

\usepackage{caption}
\usepackage{graphicx}
\usepackage{xcolor}
\usepackage{amsmath}
\usepackage{bbold}
\usepackage{booktabs} % For formal tables
\usepackage[english]{babel}
\usepackage[utf8]{inputenc}
\usepackage[T1]{fontenc}
\usepackage{times}
\usepackage{tabularx}
\usepackage{subfigure}
\usepackage{relsize}
\usepackage{spverbatim}
\usepackage{url}
\usepackage{soul}
\usepackage[subtle]{savetrees}
\usepackage[ruled]{algorithm2e} % For algorithms
\usepackage{listings}

\title{\textbf{\Large Semantic Reasoning with Differentiable Graph Transformations}}

\author{\centering
\small
Alberto Cetoli\\[0.3ex]
QBE Europe\\London, UK\\
alberto.cetoli@uk.qbe.com}

\begin{document}
\maketitle

\begin{abstract}
This paper introduces a differentiable semantic reasoner, where rules are presented as a relevant set of graph transformations. 
These rules can be written manually or inferred by a set of facts and goals presented as a training set.
While the internal representation uses embeddings in a latent space, each rule can be expressed as a set of predicates conforming to a subset of Description Logic.
\end{abstract}


\IEEEpeerreviewmaketitle

\section{Introduction}
Symbolic logic is the most powerful representation for building interpretable computational systems \cite{Garcez2020}.
In this work we adopt a subset of Description Logic \cite{Krtzsch2012ADL} to represent knowledge and build a semantic reasoner, which
derives new facts by applying a chain of transformations to the original set.

In the restricted context of this paper, knowledge can be expressed in predicate or graph form, interchangeably.
Thus, semantic reasoning can be understood as a sequence of graph transformations \cite{ehrigGraphTransformations}, which act on a subset of the original knowledge base and sequentially apply the matching rules.

In this paper, we show that rule matching can be made differentiable by representing nodes and edges as embeddings. 
After building a one-to-one correspondence between a sequence of rules and a linear algebra expression, the system can eventually train the embeddings using a convenient loss function.
The rules created in this fashion can then be applied during inference time.

Our system follows the recent revival of hybrid neuro-symbolic models \cite{Garcez2020}, combining insights from logic programming with deep learning methods. 
The main contribution of this work is to show that reasoning over graphs is a learnable task.
While the system is presented here as a proof of concept, we show that differential graph transformations can effectively learn new rules by training nodes, edges, and matching thresholds through backpropagation.

In Sec. \ref{sec:ProblemStatement} we describe in detail the fundamentals of our reasoner, with working examples shown in Sec. \ref{sec:Examples}.
Sec. \ref{sec:related_works} reviews specific connections with prior works and finally a few remarks in Sec. \ref{sec:conclusions} conclude the paper
\footnote{The relevant code can be found at \url{https://github.com/fractalego/dgt/}}.

\section{Problem statement}
\label{sec:ProblemStatement}

\begin{figure*}[t!]
\centering
    \includegraphics[width=0.95\textwidth]{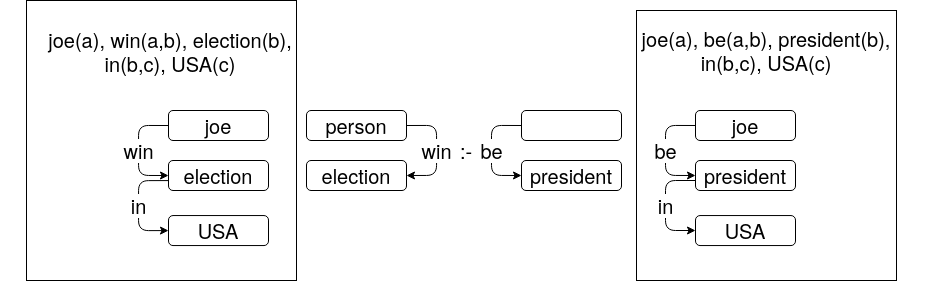}
    \caption{A matching rule example, as explained in \ref{sec:rules}.
    The facts on the left are "transformed" into the ones on the right \\following the application of the relevant rule.
    This picture makes explicit the dual nature of the predicates/graph representation. 
}
    \label{fig:single_rule}
\end{figure*}

The system presented here is a semantic reasoner inspired by the early STRIPS language \cite{STRIPS1971}. 
It creates a chain of rules that connects an initial state of \emph{facts} to a final state of inferred predicates. 
Each rule has a set of pre- and post-conditions, expressed here using a subset of Description Logic (DL). 
In the following, we restrict our DL to Assertional Axioms (ABox). 
Thus, each fact can be represented as a set of predicates, or - equivalently - as a graph with matching rules as described below.

\subsection{Rules as graph transformations}
\label{sec:rules}
We use a predicate form to represent facts, rules, and intermediate states, as shown in Fig. (\ref{fig:single_rule}).
For example the semantics for "Joe wins the election in the USA" is captured in the following form
\begin{lstlisting}{language=Java}
joe(a), win(a,b), election(b), in(b,c), USA(c)
\end{lstlisting}
In the prior example, $joe$, $election$, and $USA$ are nodes of the semantic graph, whereas $win$ and $in$ are convenient relations to represent the graph's edges.

The rules are specified with a MATCH/CREATE pair as below
\begin{lstlisting}{language=Java}
MATCH person(a), win(a,b), election(b)
CREATE (a), be(a,b), president(b)
\end{lstlisting}
The MATCH statement specifies the pre-condition that triggers the rule, while the CREATE statement acts as the effect - or post-condition - after applying the rule.
The result of applying this rule is shown in Fig. \ref{fig:single_rule}, where a new state is created from the original fact.
Notice that the name $joe$ (which matches $person$) is propagated forward to the next set of facts.
%This only happens if one of the predicates after the CREATE section is nameless.

By applying rules in sequence one builds a inferential chain of MATCH and CREATE conditions. 
After each rule the initial facts graph is changed into a new set of nodes and edges.
This chain of graph transformations builds a path in a convenient semantic space, as shown in Fig. \ref{fig:path}.
One of this paper's main result is to show that there is a one-to-one correspondence between the chain of matching rules and a chain of linear algebra operations.

\subsection{Nodes and edges as embeddings}
Both nodes and edges are represented as embeddings in a latent space.
For convenience, in the current work the vocabulary of possible nodes matches the Glove 300dim dataset \cite{pennington2014glove}, whereas edges are associated random embeddings linked to the relevant ontology.

\subsection{Nodes and edges matching}
A rule is triggered if the pre-conditions graph is a sub-isomorphism of the facts.
Each node and edge of the preconditions has a learnable threshold value $t$. 
Two items match if the dot product between their embeddings is greater than a specific threshold.
In the predicate representation, we make explicit these trainable thresholds by adding the symbol $>$ to the predicate's name. 
In this way, the rule in \ref{sec:rules} becomes

\begin{lstlisting}{language=Java}
MATCH person>0.6(a), win>0.7(a,b), election>0.6(b)
CREATE (a), be(a,b), president(b)
\end{lstlisting}
indicating that - for example - $joe$ and $person$ would only match if their normalized dot product is greater than $t=0.6$.
In the Description Logic framework this is equivalent to a \emph{individuality assertion}
\begin{equation*}
    joe \approx person \, 
       \iff \,
       \mathrm{embedding}(joe) ~
       {\Large\cdot} ~ \,\mathrm{embedding}(person) > t
\end{equation*}
Matching facts and preconditions creates a \emph{most general unifier} (MGU) that is propagated forward on the inference chain. 

\subsection{Creating a trainable path}
\begin{figure*}[t!]
    \centering
    \includegraphics[width=0.95\textwidth]{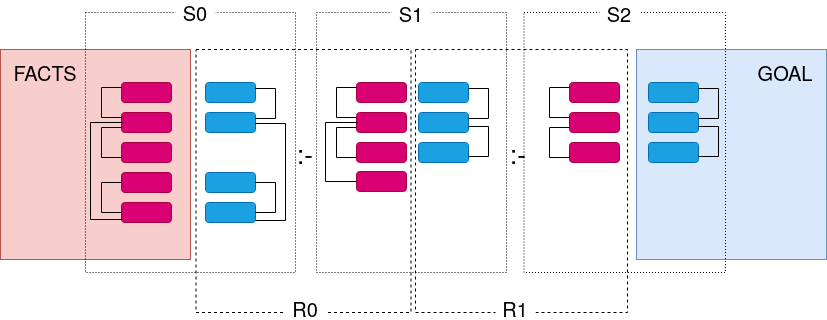}
    \caption{Example of a graph matching path.
    The facts on the left (represented as a graph of connected embeddings) are transformed through a chain of pre- and post-conditions \\into the goal on the right. This chain of rules is equivalent to a sequence of linear algebra operations, where the truth values of each predicated are propagated forward through a set \\of $S$ and $R$ matrices.
    }
    \label{fig:path}
\end{figure*}

During training, the final state is a \emph{goal} of the system, as shown in Fig. \ref{fig:path}.
The system learns how to create rules given a set of template \emph{empty rules}, where the embeddings for each node and edge are chosen randomly. 
These templates are specified prior to the training using $*$ to indicate a random embedding, as in the following
\begin{lstlisting}{language=Java}
MATCH *(a), *(a,b), *(b)
CREATE (b), *(b,d), *(d)
\end{lstlisting}

In the current state of development, the algorithm generates all possible paths - compatibly with boundary conditions - and then applies to each of them the training algorithm explained below.
A more efficient method will be pursued in future works.

\subsection{Training of the embeddings and thresholds}

At every step of the inference chain the collection of predicates changes according to the order of transformations.
At every step $i$ we employ a vector $f_i$ that signals the truth value of each predicate.
%This vector is part of the "state" of the system, together with the matrix of embeddings of each node $F_i$.
For computational reasons, the dimensions of this vector must be fixed in advance and set to the maximum size of the predicate set.
%The size $n$ of the "state" vector is then reflected onto the size of the matrices in the discussion that follow.
The first value $f_0$ is a vector of ones, as every predicate in the knowledge base is assumed to be true.

At the end of the resolution chain there is a "goal" set of predicates, usually less numerous than the initial set of facts.
A vector $g$ indicates the truth conditions of the goal predicates. 
This vector - also of size $n$ - contains a number of ones equal to the number of goal nodes and is zero otherwise.
The application of a rule can then be described by two matrices: the similarity matrix $S$ and the rule propagation matrix $R$.

A \textbf{similarity matrix} describes how well a set of facts matches the preconditions.
\begin{equation}
\label{eq:similarity}
S_i =  M_i \odot \mathrm{Softmax}(P_i^{T} \, F_i - T_i)
\end{equation}
Where $P_i$ is the matrix with the preconditions's nodes as colums, $F_i$ is the matrix with the fact nodes as columns at step $i$.
$M_i$ is the matrix of the matches, bearing value of $1$ if two nodes match and vanishing otherwise.
For example if the first node of the preconditions matches the second node of the facts, the matrix will have value $1$ at position (1, 0).

The matrix $T_i$ is a bias matrix whose columns are the list of (trainable) thresholds for each predicate in the pre-conditions $T_i = \left[ t^i_1, ~t^i_2, ~...~ t^i_n \right]$.
This bias effectively enforces the matching thresholds: 
A negative value as an argument to $\mathrm{Softmax}$ will lead to an exponentially small result after the operation.

All the matrices $M$, $P$, and $F$ are square matrices $\in \mathcal{R}^{n\,\times\,n}$.
Eq. \ref{eq:similarity} is reminiscent of self-attention \cite{vaswani_attention}, with an added bias matrix $T$ and a mask $M$.

A \textbf{rule propagation matrix} $R$ puts into contact the left side of a rule with the right side. 
The idea behind $R$ is to keep track of how information travels inside a single rule.
In this work we simplify the propagation matrix as a fully connected layer with only one trainable parameter.
For example, if the chosen size $n$ is 4, 
a rule with three pre-conditions and two post-conditions nodes has an $R$ matrix as

\begin{equation}
R_i = w\, \left[ {\begin{array}{cccc}
   1 & 1 & 1 & 0 \\
   1 & 1 & 1 & 0 \\
   0 & 0 & 0 & 0 \\
   0 & 0 & 0 & 0 \\
  \end{array} } \right]
\end{equation}
where $w$ is the "weight" of the rule.
Given a first state $f_0$, the set of truth condition after $n$ steps is
\begin{equation}
\label{eq:matrix_chain}
f_n =  S_{n-1}\,... \, R_1\,S_1\,R_0\,S_0\,f_0
\end{equation}
This final state  $f_n$ is compared against the goal's truth vector $g$ to create a loss function.

The training of the relation embeddings follows the same sequence of operations as for the nodes.
A set of truth vectors $f^r$ and states $F^r$ is acted upon the relation similarity matrix 
\begin{equation}
S_i^r = M_i^r \odot \mathrm{Softmax}({P^r_i}^{T} \, F^r_i - t^{r}_i)\,,
\end{equation}
and the corresponding rule propagation matrix for relations $R^r_i$, leading to the final truth vector for relations
\begin{equation}
\label{eq:rel_chain}
f^r_n = S^r_{n-1}\,... \, R^r_1\,S^r_1\,R^r_0\,S^r_0\,f^r_0\,.
\end{equation}
Following the example of the nodes, a goal vector for the relations is named $g^r$, containing the desired truth conditions for relations at the end of the chain.

The system learns the nodes and edges embeddings of the rules, while the initial facts and the goal are frozen during training.
The system also learns the \emph{matching thresholds} $t$ and each rule's \emph{weight} $w$.
Following Eqs. \ref{eq:matrix_chain} and \ref{eq:rel_chain}, the final loss function is 
computed as a binary cross entropy expression
\begin{equation}
\mathcal{L} = g~\mathrm{log}(f_n) 
              + g^r~\mathrm{log}(f_n^r)
\,.
\end{equation}
The system can in principle be trained over a set of multiple facts and goal pairs, in which case the loss function is the sum of all the pairs' losses. For simplicity, in this paper we limit the training to a single pair of facts and goal.

In order to avoid the Sussman anomaly the same rule can only be used once in the same path.

\section{Examples and discussion}
\label{sec:Examples}

\subsection{One-rule learning}

As a toy example we want the system to learn that if someone is married to a "first lady", then this person is president.
The facts are
\begin{lstlisting}{language=Java}
person(a), spouse(a,b), person(b), be(a,c), first-lady(c)
\end{lstlisting}

and the goal is
\begin{lstlisting}{language=Java}
person(a), profession(a,b), president(b)
\end{lstlisting}

Given the empty rule
\begin{lstlisting}{language=Java}
MATCH *(a), *(a,b), *(b), *(a,c), *(c) 
CREATE (b), *(b,d), *(d)
\end{lstlisting}

The system correctly learns the rule that connects the the facts with the goal.
\begin{lstlisting}{language=Java}
MATCH person>0.6(a), first-lady>0.6(b), person>0.6(c), 
      be>0.63631916(a,b), spouse>0.6338593(a,c) 
CREATE (b), president(d), profession(b,d)
\end{lstlisting}
While trivial, this is a fundamental test of the capacity of the system to learn the correct transformation.
The matching thresholds have been clipped and cannot go below $0.6$ in training.

While a successful result is almost guaranteed by choosing a rule that closely matches the boundary conditions, the system is proven capable of converging onto the correct embeddings and thresholds using just backpropagation.

\subsection{Chained two-rules learning}
While a single-rule transformation can be useful in a few edge cases, the real power of semantic reasoning comes from combining rules together. 
In this section we show - using another toy example - that the system can learn two rules at the same time.
The simplified task is as in the following: to learn that "if a fruit is round and is delicious, then it is an apple." 
The facts are
\begin{lstlisting}{language=Java}
fruit(a), be(a,b), round(b), be(a,c), delicious(c)
\end{lstlisting}

and the goal is
\begin{lstlisting}{language=Java}
fruit(a), be(a,b), apple(b)
\end{lstlisting}

The system is given the two template rules to fit
\begin{lstlisting}{language=Java}
MATCH *(a), *(a,b), *(b), *(a,c), *(c)
CREATE (b), and(b,c), (c) 

MATCH *(a), and(a,b), *(b) 
CREATE *(c), *(c,d), *(d) 
\end{lstlisting}
Notice the "and" relations in the templates.
These relations are frozen during training and constitute another constraint for the system to satisfy.
In the end, our model learns the correct rules
\begin{lstlisting}{language=Java}
MATCH fruit>0.6(a), round>0.6(b), delicious>0.6(c),
      be>0.6953449(a,b), be>0.6957883(a,c) 
CREATE (b), (c), and(b,c)

MATCH round>0.6(a), delicious>0.6(b), and>0.9(a,b) 
CREATE fruit(c), apple(d), be(c,d)
\end{lstlisting}
which satisfy the goal when chained.

Here we forced the system to apply two rules since no single template would fit the boundary conditions.
Of particular interest is the fact that the system learned the preconditions of the second rule $round>0.6(a), ~ delicious>0.6(b), ~ and>0.9(a,b)$.
This is not a trivial task, given that it started training with random embeddings and the only information about the correct values is the one propagated forward from the first rule.

\section{Related works}
\label{sec:related_works}

Neuro-symbolic reasoning has been an intriguing line of research in the past decades \cite{neurosymb_book2002,Garcez2009NeuralSymbolicCR}. 
Some recent results make use of a Prolog-like resolution tree as a harness where to train a neural network
\cite{rockt2017,minervini2018towards,weber-etal-2019-nlprolog,minervini2020differentiable}.
Our work is similar to theirs, but builds upon a STRIPS-like system instead of Prolog.
A different approach employs a Herbrand base for inductive logic programming in a bottom-up solver \cite{evansILP}.

Finally, one can see our method as a sequence of operations that create or destroy items sequentially. 
Each (differential) transformation brings forward a new state of the system made by discrete elements.
These types of algorithms have already been investigated in the Physics community, for example in \cite{sandvik2002}.

\section{Conclusions and future work}
In this work we presented a semantic reasoner that leverages on differential graph transformations for rule learning.
The system is build through a one-to-one correspondence between a chain of rules and a sequence of linear algebra operations. 
Given a set of facts, a goal, and a set of rules with random embeddings, the reasoner can learn new rules that satisfy the constraints.
The rules are then written as a set of predicates with pre- and post-conditions, a more interpretable representation than embeddings and weights.

The system presented here is limited in speed and - as a consequence - volume of training data. 
This is mostly due to our path-creation algorithm, which generates all possible paths given a set of rules.
A more efficient algorithm would employ a guided approach to path creation, similar to the method in \cite{minervini2020differentiable}.
A different and possibly novel efficiency gain could be found in a Monte Carlo method, where the path converges to the correct one through means of a Metropolis algorithm.
This last approach has already found application in the Computational Physics community and could be useful in our approach as well.

Another open question resides on whether the system is able to generalize, given a multiple set of facts and goals.
This last inquiry will need a faster algorithm and will be pursued in a future work.

\label{sec:conclusions}

%\section*{Acknowledgements}

\bibliographystyle{IEEEtran}
\bibliography{biblio}
\end{document}